# Concise Thoughts: Impact of Output Length on LLM Reasoning and Cost

Sania Nayab◇  Giulio Rossolini◇  Marco Simoni⋆‡  Andrea Saracino◇⋆
Giorgio Buttazzo◇  Nicolamaria Manes†  Fabrizio Giacomelli†

◇Department of Excellence in Robotics and AI, Scuola Superiore Sant'Anna, Pisa, Italy
⋆Institute of Informatics and Telematics, National Research Council of Italy
‡Sapienza Università di Roma,  †Mediavoice Srl - Roma e Napoli, Italy

## Abstract

Today's large language models (LLMs) can solve challenging question-answering tasks, and prompt engineering techniques, such as chain-of-thought (CoT), have gained attention for enhancing the explanation and correctness of outputs. However, many models and techniques tend to produce excessively verbose and lengthy answers, leading to issues with both conciseness and generation time. To address this, this paper analyzes the impact of output lengths on LLM inference pipelines by introducing and proposing novel metrics to evaluate the *correct conciseness* of a model and related prompting techniques. Then, we examine the impact of controlling output length through a refined prompt engineering strategy, Constrained-CoT (CCoT), which encourages the model to produce more concise outputs. To better understand the effects of such a prompt, we also introduce two additional scores for analyzing the conciseness, measured in terms of redundancy and information flow in generated answers. Experiments on pretrained LLMs and multiple datasets demonstrate the benefits of the proposed metrics and the effectiveness of CCoT across different models.

## 1 Introduction

In recent years, large language models (LLMs) have demonstrated remarkable capabilities in tackling complex question-answering tasks, making significant strides in natural language understanding and generative AI (Taori et al., 2023; Chiang et al., 2023; Dolly, 2023; Geng et al., 2023). The continuous advancements made in architectures and training methods played a crucial role in enhancing the performance of these models. Alongside these developments, prompt techniques have also seen substantial evolution. One such technique that has attracted considerable attention is chain-of-thought (CoT) prompting (Wei et al., 2022; Fu et al., 2023), which enhances the explanation and correctness of the output by encouraging the LLM to articulate its answer through intermediate reasoning steps.

Despite its advantages, the CoT prompting can lead to long outputs, increasing the time required for the model to generate a response. This is due to the nature of autoregressive transformers, which decode text word by word (Vaswani et al., 2017; Shekhar et al., 2024), which implies that the time required to generate a response is unbounded and heavily influenced by the length of the reasoning provided, as demonstrated in Section 3. Such lengthy and variable delays in responses can be undesirable when an LLM interacts with a user in an interactive conversation. Furthermore, especially for complex models, long answers imply a loss of conciseness and even of precision in the answer, with a performance degradation which is not only bound to computation time. This issue highlights the need to consider *(i)* metrics for evaluating the conciseness of the outputs and *(ii)* solutions to avoid excessively long chains of reasoning.

To address this, the first part of this work emphasizes the importance of accounting for the length of an answer in its correctness evaluation, as an indicator of computational cost. This is achieved by introducing three novel metrics (HCA, SCA, and CCA) that assess both the brevity and correctness of a generated answer. The proposed metrics aim to reweight the accuracy of a model by considering aspects related to output length that impact inference time and time predictability.

Then, to address the significant increase in output length caused by CoT techniques, the second part of this work explores how to leverage the benefits of CoT advances while getting control over the length of CoT reasoning through specific prompt designs. To this end, we introduce a refined prompt strategy called Constrained-CoT (CCoT), which encourages LLMs to generate concise outputs by explicitly limiting the reasoning length. The ap-proach allows users to set a flexible length constraint that serves as a tunable parameter, balancing the strictness of brevity in the answers. The objective is to enable controlled reasoning, ensuring that outputs are concise and computationally efficient without sacrificing accuracy.

To better assess the ability of LLMs to follow such instructions and gain deeper insights into conciseness, we also introduce additional scores that analyze the level of redundancy and the information flow in the generated answers. These scores help demonstrate, through experimental analysis, that large models (such as Llama2-70b and Falcon-40b) can effectively leverage the proposed prompt to produce more concise yet accurate responses, while still retaining useful information. This allows achieving an enhanced trade-off between accuracy and brevity, measured comprehensively using the proposed metrics. For instance, using LLaMA2 on three datasets (GSM8K, SVAMP, ASDIV), constraining the reasoning length to 30 words (CCoT-30) increases the average accuracy by 4.41% and reduces computational costs by 5.12s. These accuracy and cost improvements are better remarked and unified in the proposed metrics, showing an improvement of 10%, 9% and 9% for HCA, SCA, and CCA, respectively.

To summarize, this work provides the following main contributions:

- Introduction of the concept of Constrained-CoT (CCoT), a prompt engineering strategy designed to limit the length of answers generated by LLMs, thereby enhancing the trade-off between generation time and correct conciseness.

- Three novel metrics to evaluate the correctness of LLM outputs while accounting for the conciseness (HCA, SCA and CCA).

- Introduction of an analysis of the conciseness in terms of redundancy and information flow for a given answer, thus offering an understanding of the effects of constraining output length.

- We conducted multiple experiments to analyze the impact of CCoT across different datasets and LLMs, demonstrating its benefits in terms of inference time, accuracy and conciseness with respect to the original CoT. Furthermore, we show the benefits of adopting the proposed metrics and scores in terms of conciseness.

The rest of the paper is organized as follows: Section 2 discusses the literature related to this work; Section 3 motivates the addressed study; Section 4 presents a set of metrics that account for conciseness; Section 5 introduces the proposed CCoT approach; Section 6 introduces analysis for evaluating the conciseness of a given answer; Section 7 reports the results of a set of experiments carried out on pre-trained models with three arithmetic reasoning datasets; and Section 8 states the conclusions and discusses some future directions.

## 2 Related work

To the best of our knowledge, most recent works on LLMs focused on increasing their accuracy (Jiang et al., 2020; Kaplan et al., 2020; Zhu et al., 2023). However, as models scale up, they tend to generate more extensive and articulated responses (Bhargava et al., 2023), which can introduce other problems, such as hallucinations (where the model produces information that appears plausible but not grounded (Kadavath et al., 2022), or unnecessarily long explanations (Qiu et al., 2024; Azaria and Mitchell, 2023)), which can obscure key information, making it difficult for users to extract relevant content efficiently (Khashabi et al., 2021; Wang et al., 2024b). To filter out useless reasoning, Li et al. (2021) proposed a multi-hop processing technique, where an extraction task on the encoder to obtain the rationale for an answer, which is the most relevant piece of text in an input prompt to a given question.

To further improve the accuracy of LLMs, several prompt engineering approaches have been presented in recent years (Qin and Eisner, 2021). Prompt engineering involves the strategic design of input patterns to guide the model toward generating more accurate and relevant responses (Reynolds and McDonell, 2021; Marvin et al., 2023). However, most of these approaches have been conceived to enhance model accuracy, increasing the output length. For instance, Lo (2023) and Strobelt et al. (2022) introduced prompt-based approaches by adding task-specific patterns to frame the input data. While these methods allow boosting accuracy, they can also produce longer outputs due to the additional context and detail introduced by the prompt, making it challenging to provide factual and concise answers (Shi et al., 2023).

Another form of prompt engineering was proposed to improve reasoning within the conclusive answer. In this context, Chain-of-Thought (CoT) prompting (Wei et al., 2022) is one of the most notable methods, showing significant benefits in QA tasks by requiring the model to provide a step-by-step explanation along with the final response. However, as also highlighted in Section 3, answers generated with CoT tend to be lengthy, hence increasing the generation time (Liu et al., 2018; Takase and Okazaki, 2019).

Given the substantial amount of work focused on improving the accuracy of LLMs, it is not surprising that most of the adopted metrics (Lin, 2004; Stallings and Gillmore, 1971) and benchmarks (Clark et al., 2018; Lin et al., 2021) only address the correctness of the responses, without paying attention to conciseness and response times (Bhargava et al., 2023; Chiang and Lee, 2024). In other tasks too, such as reasoning in control engineering, the focus has been primarily on correctness rather than consistency or conciseness (Kevian et al., 2024). Additionally, several studies have addressed computational cost challenges but not the conciseness. For example, Wang et al. (2024a) evaluated the budget-aware reasoning capabilities of LLMs, while Zheng et al. (2024b) proposed an inference pipeline to improve processing speed. Other works have explored similar optimization approaches, including (Hao et al., 2024) and (Bi et al., 2020). In addition, Chiang and Lee (2024) proposed a benchmark to study LLMs accuracy while incorporating a manual redundancy assessment.

Despite recent advancements, several key aspects remain not sufficiently explored: *(i)* the impact of concise answers on inference cost and time predictability; *(ii)* the integration of such aspects into unified metrics that evaluate LLMs not only in terms of correctness but also conciseness; and *(iii)* an understanding of the analysis of conciseness through the conciseness based on the embeddings content extracted by the generated answers.

**This work.** To address these challenges, this work introduces novel metrics that jointly account for the conciseness and correctness of generated responses. Additionally, two new scores are proposed to assess conciseness from the embeddings produced by the model. These scores focus on the importance of reasoning steps by analyzing redundancy and information flow. Then, to evaluate the ability of LLMs to control the length of reasoning in their outputs, this work introduces a refined version of the CoT prompting (Wei et al., 2022), termed Constrained Chain-of-Thought (CCoT). This approach explicitly guides the model to limit the length of its reasoning while preserving the quality of its answers and improving the inference time. This is achieved by improving the conciseness of the responses, which is analyzed and assessed using the proposed scores.

## 3 Motivations

The output generation time of an LLM depends on various factors, including the model architecture, the pre-and post-processing steps, the answer decoding process, and the question posed, also considering the use of prompt engineering approaches. While the computational cost due to the architecture is well understood, the influence of the other aspects on the overall generation time is less clear and requires further investigation. More formally, an LLM can be represented as a function $f$ that takes as input a prompt $x$ with $\mathcal{N}(x)$ tokens[1] and generates an output $\hat{y} = f(x)$, having $\mathcal{N}(\hat{y})$ tokens, where $\mathcal{N}$ is a length operator that simply counts the number of tokens. The input $x$ can be considered as composed of the original user input $x_{us}$ and a prompt engineering text $x_p$, depending on the technique used. For instance, in a zero-shot CoT setting, the prompt can be computed as $x = concat(x_{us}, x_p)$, where $x_p$ is an explicit request for providing reasoning steps in the answer and $concat(a, b)$ is the concatenation operator that merges two vectors $a$ and $b$ into a single one.

In an encoder-decoder architecture, as the one used by Transformers (Vaswani et al., 2017), let $f_e(x)$ and $f_d(x)$ denote the functions associated with the encoder and the decoder, respectively. Then, the output $\hat{y}$ is a list of tokens $[a^{(1)}, \ldots, a^{(\mathcal{N}(\hat{y}))}]$, where each $a^{(i)}$ is computed based on the previously generated tokens and the encoder's embedding representation $f_e(x)$. That is,

$$a^{(i)} = f_d(f_e(x), [a^{(0)}, \ldots, a^{(i-1)}]), \quad i > 0. \quad (1)$$

From Equation (1), it is clear that the larger the set of output tokens in the answer, the higher the time the model takes to generate the answer due to the increased number of times the decoder is invoked. The same consideration could also be applied to decoder only, where the model runs a new inference for each produced word.

---
[1] Even though 'tokens' and 'words' refer to different items in the sentence, for simplicity, in this work we will refer to both indistinguishably.

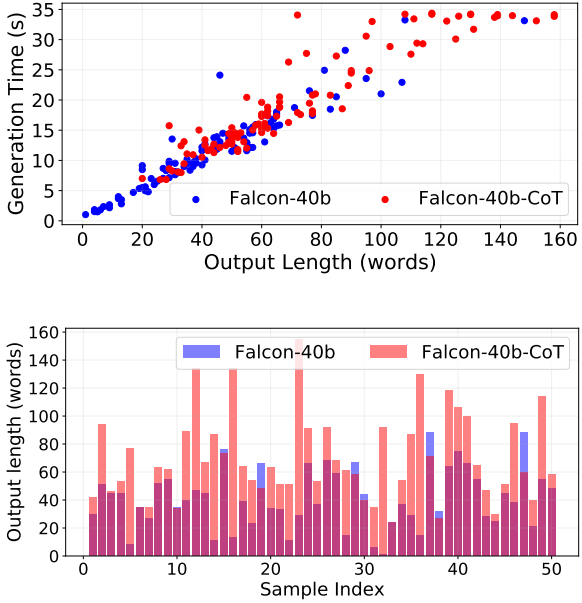

Figure 1: Analysis of the impact of CoT on Falcon-40b efficiency: (top) Relation between response time and output length, without CoT (blue dots) and with CoT (red dots), across 100 questions from the GSM8K test set. (bottom) Output words variation between output length with CoT and without CoT using 50 random samples from the GSM8K test set.

To highlight such a dependency, we conducted preliminary tests on Falcon-40B to evaluate the impact of the CoT method in answering arithmetic questions, using a subset of 100 random questions from the GSM8K dataset (Cobbe et al., 2021). The results of this test are illustrated in Figure 1a, where red and blue dots refer to answers given with and without CoT, respectively. The scatter plot shows that CoT significantly increases the output length and generation time. This suggests that while CoT improves the correctness of responses (see Section 7), more attention should be given to the time cost it introduces. To better appreciate the impact of CoT on the output length, Figure 1b reports the output length (in terms of number of generated words) produced by Falcon-40b on a set of 50 questions from GSM8K without CoT (blue bars) and with CoT (pink bars). Note that purple areas denote the areas where the two bars overlap.

## 4 Metrics for Correct Conciseness

Motivated by the previous considerations, this section presents three novel metrics to evaluate the capability of an LLM to provide *correct* as well as *concise* responses. The idea is to redefine the classic accuracy metric to integrate conciseness aspects, which as highlighted in Section 7.3 impacts significantly the generation time and the computational cost, into the LLM output's correctness. Formally, an answer $\hat{y}$ is considered correct if the conclusion extracted through a post-processing function $\Gamma$ matches the given ground truth $y$. Thus, the accuracy of an LLM can be computed as

$$\mathcal{A} = \frac{1}{N} \sum_{i=1}^{N} \mathbb{1}(\Gamma(\hat{y}), y), \qquad (2)$$

where $N$ is the number of tested samples and $\mathbb{1}(u, v)$ is the indicator function that returns 1 if $u = v$, 0 otherwise. Please note that $\Gamma$ represents a user-defined function that can be implemented based on a regular expression (e.g., by extracting specific patterns from the sentence (Fu et al., 2023)) or using pseudo-judge approaches (e.g., by using a secondary large model as a judge (Zheng et al., 2024a)).

Starting from Equation (2), the conciseness of an output $\hat{y}_i$ can be integrated with its correctness by multiplying the indicator function by a penalty term $p(\hat{y}_i) \in [0, 1]$ that decreases its value for long outputs:

$$\frac{1}{N} \sum_{i=1}^{N} \left[ \mathbb{1}(\Gamma(\hat{y}_i), y_i) \cdot p(\hat{y}_i) \right]. \qquad (3)$$

The following defines three specific metrics by setting a proper penalty function.

**Hard-$k$ Concise Accuracy:** HCA($k$). It measures the fraction of correct outputs that do not exceed a user-specified length $k$:

$$\text{HCA}(k) = \frac{1}{N} \sum_{i=1}^{N} \left[ \mathbb{1}(\Gamma(\hat{y}_i), y_i) \cdot p_{hard}(\hat{y}_i, k) \right],$$

where

$$p_{hard}(\hat{y}_i, k) = \begin{cases} 1 & \text{if } \mathcal{N}(\hat{y}_i) \leq k \\ 0 & \text{otherwise.} \end{cases} \qquad (4)$$

This metric does not account for responses that exceed the specified maximum length, thereby promoting conciseness. We believe it could be particularly useful in scenarios where strict adherence to length constraints is essential, such as in real-time systems or environments with limited computational resources.

**Soft-$k$ Concise Accuracy:** $\text{SCA}(k, \alpha)$. It generalizes the previous metric by penalizing the correct answers that exceed the maximum length $k$ with a term that decreases exponentially with a decay factor $\alpha$:

$$\text{SCA}(k, \alpha) = \frac{1}{N} \sum_{i=1}^{N} \left[ \mathbb{1}(\Gamma(\hat{y}_i), y_i) \cdot p_{soft}(\hat{y}_i, k, \alpha) \right],$$

where

$$p_{soft}(\hat{y}_i, k, \alpha) = \min\left(1, e^{\frac{k-\mathcal{N}(\hat{y}_i)}{\alpha}}\right). \quad (5)$$

In the formula, the user-defined decay $\alpha \geq 0$ can be considered a sort of tolerance that controls how much the length impacts the overall accuracy; the higher the value of $\alpha$, the higher the tolerance for answers exceeding the specified length $k$. Note that for $\alpha = 0$, $\text{SCA}(k, 0)$ reduces to $\text{HCA}(k)$.

**Consistent Concise Accuracy:** $CCA(k, \alpha, \beta)$. It further generalizes the previous metrics by also accounting for the variation in the lengths among all the outputs obtained:

$$CCA(k, \alpha, \beta) = SCA(k, \alpha) \cdot p_{var}(\sigma, \beta)$$

where

$$p_{var}(\sigma, \beta) = \min\left(1, e^{\frac{\beta-\sigma}{\beta}}\right). \quad (6)$$

In Equation (6), $\sigma$ denotes the standard deviation of the output length distribution, whereas $\beta$ is a parameter that controls the tolerance for having large length variations: the higher the value of $\beta$, the higher the tolerance. Note that, given a tolerance $\beta$, $p_{var}(\sigma, \beta) = 1$ for $\sigma \leq \beta$, while it decreases exponentially for $\sigma > \beta$.

The CCA metric aims to promote consistency in the lengths of responses, representing an important measure when the predictability of inference time is a crucial property for a specific application. A low standard deviation $\sigma$ indicates that the model produces responses of uniform length. In contrast, a high value of $\sigma$ denotes a model with a large response variability, making predicting its timing response time difficult.

## 5 CCoT Prompting

From the results presented in Section 3, it is clear that the relationship between output length and inference time necessitates deeper awareness. To this end, this section focuses on improving the use of CoT, aiming to preserve the benefits of this technique while paying more attention to the length of the answers. This help achieve a better trade-off between efficiency and accuracy.

For this purpose, we introduce a constrained chain of thought (CCoT) prompt, which includes an explicit sentence to constrain the generated output to a maximum number of words, encouraging the model to compress its reasoning and produce a more concise answer in a reduced amount of time. As explained in Section 3, CoT-prompt can be computed as $x = concat(x_{\text{us}}, x_p)$, where $x_p$ is an explicit request for providing reasoning steps in the generated answer (e.g., *"let's think step by step"*). Technically, to push LLMs to return more concise reasoning, the CCoT-prompt is formalized as $x = concat(x_{\text{us}}, x_p, x_l)$, where $x_l$ represents the sentence that specifies the constraint on the output length (e.g., *"and limit the length of the answer to 30 words"*). Figure 2 shows an example that illustrates the difference between a CoT and a CCoT prompt. Note that the answer generated for that specific question using a CoT prompt consists of 67 words in 17.64s, while the answer generated on the same question provided with a CCoT prompt (specifying a constraint of 45 words) consists of 34 words in 11.65s, and it is still correct.

The analysis presented in the following section and experimental part in Section 7 provide a detailed evaluation of the CCoT prompting technique using various metrics and demonstrate its benefits on the conciseness along with redundancy and information flow scores of the generated answers.

## 6 Analysis of the conciseness

To provide a deeper understanding and more comprehensive comparisons of answers in terms of conciseness, we define some conciseness properties for a given answer in the following. Intuitively, we assume that the conciseness of an answer in the CoT paradigm could be derived from the following two concepts: *how many steps are required to produce a response* and *how much information is repeated in successive steps*. While the first property is easy to derive by immediately looking at the number of steps, assuming to have a similar number of steps from two answers, extracting a numerical interpretation of the conciseness could be more complicated. For this purpose, we define two additional scores: *Redundancy*, which measures

> CoT → output length 67 words; **inference time 17.64s**
>
> > Q: Tracy used a piece of wire 4 feet long to support tomato plants in the garden. The wire was cut into pieces 6 inches long. How many pieces did she obtain? **Let's think a bit step by step**
> >
> > A: First, we need to convert 4 feet to inches. There are 12 inches in a foot, so 4 feet equals 4 × 12 = 48 inches. Next, we can divide the total number of inches by the length of each piece (6 inches) to find the total number of pieces: 48 inches ÷ 6 inches = 8 pieces. Therefore, Tracy obtained 8 pieces of wire.
>
> CCoT-45 → output length 34 words; **inference time 11.65s**
>
> > Q: Tracy used a piece of wire 4 feet long to support tomato plants in the garden. The wire was cut into pieces 6 inches long. How many pieces did she obtain? **Let's think a bit step by step and limit the answer length to 45 words.**
> >
> > A: 1. Convert 4 feet to inches: 4 × 12 = 48 inches 2. Divide 48 inches by 6 inches per piece: 48 ÷ 6 = 8 pieces. So, Tracy obtained 8 pieces of wire.

Figure 2: Example of CCoT in a zero-shot setting for a question extracted from the GSM8K dataset. In each box, we show the prompt and answer provided by the LLM. The box above shows the classic CoT zero-shot approach, where long reasoning is returned as output. The box below shows the use of CCoT, which reduces the number of output words while maintaining a correct answer. In the box title, we also show the word count along with inference time to clarify the improvements of the approach.

the extent to which individual steps in a generated answer repeat or overlap in *synthetical* content, and *Information Flow*, which measures how much the *semantic* information of the previous step is repeated in the current step. To compute these two scores, both CCoT and CoT answers are divided into discrete steps, based on sentence tokenization[2].

### 6.1 Redundancy Score

The redundancy score for each step represents how much its content overlaps with other steps. Let a generated answer $\hat{y}$ be divided into $n$ steps $S = \{s_1, s_2, \ldots, s_n\}$, where each $s_i$ represents a single step (sentence). The redundancy mean score $RMS(\hat{y})$ is computed as:

$$RMS(\hat{y}) = \frac{1}{n \cdot (n-1)} \sum_i^n \sum_{j \neq i}^n \text{SyS}(s_i, s_j), \quad (7)$$

where $\text{SyS}(s_i, s_j)$ measures the syntactical similarity between steps $s_i$ and $s_j$ as the ratio of matching subsequences. This is computed as $\text{SyS}(s_i, s_j) = \frac{\text{LMS}(s_i, s_j)}{\text{TLCS}(s_i, s_j)}$, where LMS states for *Length of Matching Subsequences* and TLCS states for *Total Length of Compared Sequences* (Ratcliff et al., 1988).

These scores help to evaluate whether the reasoning process is concise or contains redundant synthetical repetition. We assume that for not concise answers we have a high redundancy, which indicates that steps are overly reliant on repeating information rather than advancing the reasoning.

[2] https://www.nltk.org/api/nltk.tokenize.sent_tokenize.html

### 6.2 Information Flow

The *Information Flow* $\mathcal{I}(s_i, s_{i+1})$ between two consecutive steps $s_i$ and $s_{i+1}$ measures the extent to which the content of $s_i$, in a semantic sense, persists in $s_{i+1}$. This can be expressed as:

$$\mathcal{I}(s_i, s_{i+1}) = SeS(s_{i+1}, s_i), \quad (8)$$

where $SeS(s_{i+1}, s_i)$ represents the semantic similarity between $s_i$ and $s_{i+1}$. In our implementation, we used *BERTScore* (Zhang et al., 2019; Devlin et al., 2018) to compute the semantic similarity between steps, which leverages contextual embeddings generated by BERT. For the last step, since there is no subsequent step, we assume $\mathcal{I}(s_n) = 0$.

The rationale behind this score is that, when the number of steps is similar, the information flow across the steps tends to be higher in less concise answers because each step relies on a more similar set of details and context from the previous one, increasing the risk of repetitiveness.

## 7 Experiments

This section presents a set of experiments carried out to evaluate the effectiveness of the proposed CCoT approach under classic metrics, as well as illustrate the benefits of the proposed metrics and scores in highlighting the trade-off between accuracy and computational cost. Specifically, the following research questions are investigated in the next experiments: RQ1 Is the CCoT approach beneficial in terms of efficiency and accuracy compared to classic CoT?; RQ2 Are the proposed metrics representative of showing this trade-off?; and RQ3 Are the CCoT answers actually more concise, and to what extent are LLMs capable of controlling the output length based on an explicit prompt request?

## 7.1 Experimental setup

All the experiments have been carried out with the Text Generation Inference (TGI) platform[3] on 8 NVIDIA A100 GPUs. Specifically, we evaluated large and open source pre-trained LLMs from Hugging Face[4], such as instruction-tuned models Falcon-40b-instruct and model trained, reinforced by utilizing private data, namely Llama2-70b-chat-hf (Touvron et al., 2023). Additionally, further experiments on smaller models were conducted to evaluate their capability in handling CCoT, with detailed results provided in Appendix Section B.

The experiments utilized three arithmetic reasoning datasets: GSM8k (Cobbe et al., 2021), SVAMP (Patel et al., 2021), and ASDIV (Miao et al., 2021), commonly used to assess models' mathematical inference and computational reasoning abilities. The GSM8k test set includes over 1.3k problems out of 8,000. The SVAMP test set contains 300 examples, and the ASDIV test set comprises 1.22k examples.

For all experiments, the effectiveness of CCoT was compared by assessing the selected LLMs both with and without CoT (base mode).

## 7.2 Impact of CCoT on Accuracy and Efficiency

This experiment was carried out to evaluate the impact of *CCoT* on computation time and accuracy (RQ1). In particular, the selected LLMs were evaluated on three datasets using plain prompt (base), *CoT*, and *CCoT* with different length constraints:15, 30, 45, 60, 100. The results are presented in Figure 3 for Llama2-70b (top) and for Falcon-40b (bottom), showing accuracy and generation time in the first and second rows, respectively.

[3]https://huggingface.co/docs/text-generation-inference
[4]https://huggingface.co/blog/os-llms

Considering the results with Llama2-70b, CCoT prompting demonstrates the ability to reduce generation time compared to CoT and, in most cases, achieves a time reduction similar to or better than plain prompting (base). Additionally, it generally improves or minimally impacts accuracy, thereby enhancing the trade-off in both directions. For instance, the average generation time decreases from 30.09 seconds with CoT to a maximum of 23.86 seconds with CCoT on GSM8K with Llama2-70b, achieved with a length constraint of 100, and further reduces with stricter constraints. At the same time, also the accuracy consistently improves, for example, with the GSM8k dataset, the accuracy of Llama2-70b increases from 36% with CoT to 37% (with CCoT-30) and 41.77% (with CCoT-100).

Similar observations can be made for the results with Falcon-40b (bottom part), where the CCoT approach improves the trade-off between efficiency and accuracy across all three datasets. In particular, CCoT achieves better accuracy overall for SVAMP and ASDIV, while significantly reducing generation time. This improvement in terms of accuracy, however, does not occur for the GSM8K dataset, where the sentences are more complex than those in the other two datasets. We believe that such a complexity makes it more challenging for a medium-sized model like Falcon-40b to effectively handle the CCoT constraint. Nonetheless, the accuracy remains higher than the base mode, indicating that the CoT-based approach still provides benefits.

We also acknowledge that different behaviors may arise when dealing with datasets of varying complexity, as the nature of the questions can differ. For instance, CCoT-15 outperforms other CCoT variations because its 15-word responses potentially align better with the simpler nature of this dataset. We argue that this effect is closely related to the complexity of the questions, which makes the constraint impactful even on accuracy.

## 7.3 Analysis of the *correct conciseness*

Based on the previous results, CCoT demonstrates clear benefits in balancing accuracy and computational efficiency. To unify these aspects, we introduced new metrics in Section 4 that integrate answer length with correctness. The analysis presented in Tables 1 and 2 for Llama2-70b and Falcon-40b, respectively, highlights how these metrics effectively combine cost and accuracy (RQ2), while also remarking the advantages of CCoT.

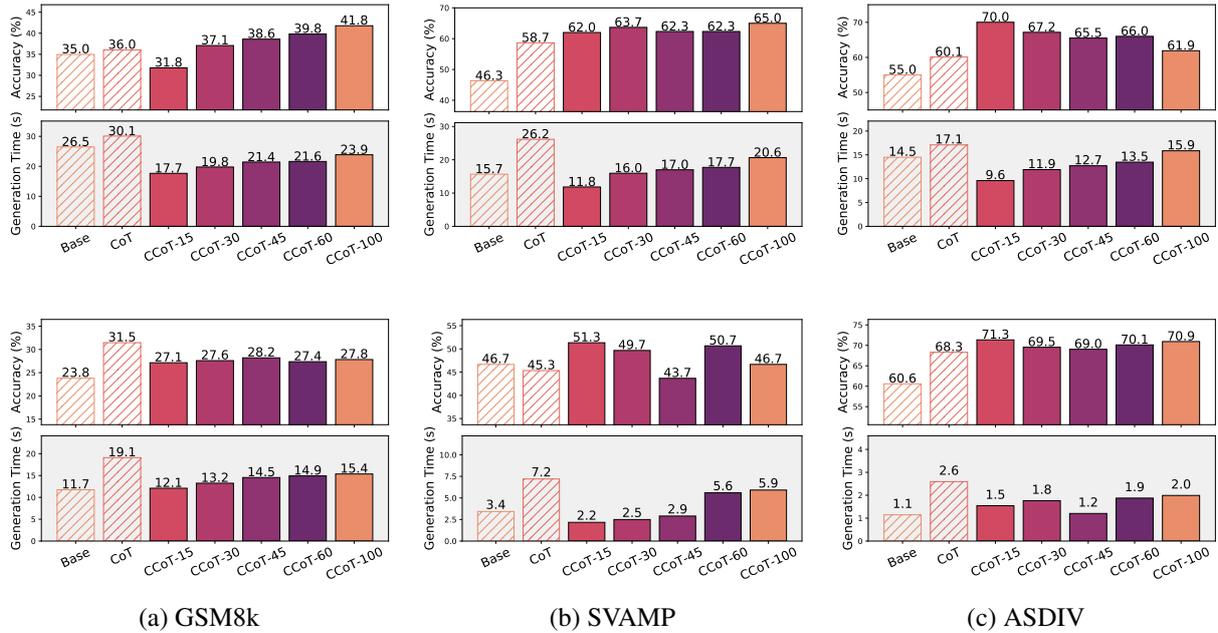

(a) GSM8k  (b) SVAMP  (c) ASDIV

Figure 3: Accuracy (white plots, the higher the better) and mean generation time (background colored plots, the lower the better) for Llama2-70b (top row) and Falcon-40b (bottom row) on the GSM8K, SVAMP, and ASDIV datasets. The models are evaluated using plain prompts (base), CoT, and CCoT under different len-constraints.

**HCA evaluation.** The *Hard-k concise accuracy* evaluates the accuracy considering only the correct answers whose length is less than a specified value $k$. The top parts of Tables 1 and 2 report the value of this performance index across the three datasets, when using the different prompt approaches and for different values of $k$.

Specifically, for both Llama2-70b and Falcon-40b, the use of CCoT gets better results compared to base and CoT prompts across all values of $k$. Notably, for lower values of $k$, CoT prompts exhibit a significant reduction in performance, while this accuracy drop can be mitigated by using CCoT with strict length constraints, such as 15 or 30.

**SCA evaluation.** We also evaluated both models using the *Soft Conciseness Accuracy (SCA)*, across different $k$ values and $\alpha$, where $\alpha$ represents a tolerance for accepting answers longer than the desired limit $k$. This metric is a generalization of the *HCA*, giving more flexibility in considering correct answers that are larger but still close to the desired length $k$.

The SCA values computed for Llama2-70b and Falcon-40b on the datasets are reported in center parts of the tables for different values of $k$ and a fixed tolerance value $\alpha = 10$. For both models, the SCA values in CCoT settings are often comparable to HCA values for high values of $k$, such as 80 or 100. This is because, as discussed in Sec. 7.5, for such lengths, the CCoT prompts are effective at returning outputs below the desired limit, making the tolerance less necessary. Conversely, for smaller $k$ values, such as $k = 40$, SCA starts exceeding HCA, indicating that some correct answers have a length larger than $k$. However, for such $k$ values of, using a tolerance $\alpha$ results in more pronounced improvements for CCoT prompts compared to Base and CoT. This means that, although many correct outputs are longer than $k$, under CCoT the model is still encouraged to constrain them close to $k$, thus achieving a higher score. This effect is particularly noticeable on Llama2-70b, which is more capable of controlling the length and produce correct outputs than Falcon-40b.

**CCA evaluation.** The *Consistent Concise Accuracy* measures the capability of a model to generate correct answers whose lengths do not vary significantly, and therefore are consistent with the specified constraint. The CCA requires a third parameter $\beta$ (in addition to $k$ and $\alpha$), denoting a tolerance on the output length variability. The bottom parts of the tables report the $CCA$ scores obtained on Llama2-70b and Falcon-40b for $\alpha = 10$, $\beta = 40$, and different values of $k$, for the various prompting methods. According to these settings, the CCoTs results in a clear improvement compared to CoT prompting and base, for both Llama2-70b and Falcon-40b. However, for high CCoT length constraints (e.g., 100), the CCA score tends to decrease, which does not happen with the other two metrics. This can be explained by considering that an increased length constraint gives the model more freedom to generate outputs with higher variations, as discussed in Section 7.5.

|  | Base | CoT | CCoT15 | CCoT30 | CCoT45 | CCoT60 | CCoT100 |
|---|---|---|---|---|---|---|---|
| **GSM8K - HCA** | | | | | | | |
| H-∞ | 35.0 | 36.0 | 31.8 | 37.1 | 38.6 | 39.8 | **41.8** |
| H-100 | 29.9 | 22.9 | 31.2 | 35.3 | 37.5 | 38.7 | **38.9** |
| H-80 | 22.0 | 15.4 | 29.2 | 31.8 | 33.1 | **35.0** | 31.6 |
| H-40 | 4.8 | 0.8 | **12.7** | 10.8 | 8.0 | 8.5 | 4.5 |
| **SVAMP - HCA** | | | | | | | |
| H-∞ | 46.3 | 58.7 | 62.0 | 63.7 | 62.3 | 62.3 | **65.0** |
| H-100 | 46.3 | 50.0 | 59.7 | 61.0 | 61.0 | 59.7 | **61.7** |
| H-80 | 46.3 | 41.0 | 56.7 | **57.7** | 57.0 | 54.3 | 54.7 |
| H-40 | 23.3 | 12.0 | **44.7** | 34.0 | 30.0 | 30.0 | 20.0 |
| **ASDIV - HCA** | | | | | | | |
| H-∞ | 52.8 | 60.3 | **67.2** | 65.6 | 64.3 | 65.0 | 61.4 |
| H-100 | 52.8 | 60.3 | **67.2** | 65.6 | 64.3 | 65.0 | 61.4 |
| H-80 | 52.8 | 60.3 | **67.2** | 65.6 | 64.3 | 65.0 | 61.4 |
| H-40 | 31.7 | 23.6 | **55.5** | 46.4 | 44.4 | 43.8 | 29.5 |
| **GSM8K - SCA ($\alpha = 10$)** | | | | | | | |
| SCA-100 | 31.4 | 26.9 | 31.5 | 36.3 | 38.1 | 39.2 | **40.1** |
| SCA-80 | 25.6 | 19.0 | 30.1 | 33.8 | 35.3 | **36.6** | 35.0 |
| SCA-40 | 8.5 | 3.3 | **18.0** | 16.9 | 15.4 | 16.3 | 11.0 |
| **SVAMP - SCA ($\alpha = 10$)** | | | | | | | |
| SCA-100 | 46.3 | 52.3 | 60.3 | 61.6 | 61.3 | 60.6 | **62.7** |
| SCA-80 | 46.3 | 45.6 | 58.4 | **59.2** | 59.1 | 57.6 | 58.1 |
| SCA-40 | 31.4 | 19.5 | **48.2** | 41.1 | 39.6 | 38.7 | 31.6 |
| **ASDIV - SCA ($\alpha = 10$)** | | | | | | | |
| SCA-100 | 52.8 | 60.3 | **67.2** | 65.6 | 64.3 | 65.0 | 61.4 |
| SCA-80 | 52.8 | 60.3 | **67.2** | 65.6 | 64.3 | 65.0 | 61.4 |
| SCA-40 | 39.5 | 35.2 | **60.4** | 54.0 | 52.1 | 51.7 | 40.9 |
| **GSM8K - CCA ($\alpha = 10, \beta = 40$)** | | | | | | | |
| CCA-100 | 31.4 | 26.9 | 31.5 | 36.3 | 38.1 | 39.2 | **40.1** |
| CCA-80 | 25.6 | 19.0 | 30.1 | 33.8 | 35.3 | **36.6** | 35.0 |
| CCA-40 | 8.5 | 3.3 | **18.0** | 16.9 | 15.4 | 16.3 | 11.0 |
| **SVAMP - CCA ($\alpha = 10, \beta = 40$)** | | | | | | | |
| CCA-100 | 46.3 | 52.3 | 60.3 | 61.6 | 61.3 | 60.6 | **62.7** |
| CCA-80 | 46.3 | 45.6 | 58.4 | **59.2** | 59.1 | 57.6 | 58.1 |
| CCA-40 | 31.4 | 19.5 | **48.2** | 41.1 | 39.6 | 38.7 | 31.6 |
| **ASDIV - CCA ($\alpha = 10, \beta = 40$)** | | | | | | | |
| CCA-100 | 52.8 | 60.3 | **67.2** | 65.6 | 64.3 | 65.0 | 61.4 |
| CCA-80 | 52.8 | 60.3 | **67.2** | 65.6 | 64.3 | 65.0 | 61.4 |
| CCA-40 | 39.5 | 35.2 | **60.4** | 54.0 | 52.1 | 51.7 | 40.9 |

Table 1: Llama2-70b

|  | Base | CoT | CCoT15 | CCoT30 | CCoT45 | CCoT60 | CCoT100 |
|---|---|---|---|---|---|---|---|
| **GSM8K - HCA** | | | | | | | |
| H-∞ | 23.8 | 31.5 | 27.1 | 27.6 | **28.2** | 27.4 | 27.8 |
| H-100 | 23.7 | **29.3** | 26.7 | 27.3 | 27.2 | 26.8 | 27.4 |
| H-80 | 22.8 | **26.8** | 25.8 | 26.1 | 25.9 | 25.2 | 26.0 |
| H-40 | 13.0 | 10.4 | **13.4** | **13.4** | 12.2 | 11.8 | 12.5 |
| **SVAMP - HCA** | | | | | | | |
| H-∞ | 46.7 | 45.3 | **51.3** | 49.7 | 43.7 | 50.7 | 46.7 |
| H-100 | 46.7 | 42.3 | **51.3** | 49.3 | 43.0 | 48.3 | 45.3 |
| H-80 | 46.3 | 39.7 | **49.7** | 47.3 | 40.7 | 46.7 | 44.3 |
| H-40 | **38.7** | 20.3 | 36.7 | 36.0 | 31.7 | 33.0 | 32.7 |
| **ASDIV - HCA** | | | | | | | |
| H-∞ | 60.6 | 68.3 | **71.3** | 69.5 | 69.0 | 70.1 | 70.9 |
| H-100 | 60.6 | 67.1 | **71.2** | 69.1 | 68.8 | 69.7 | 70.1 |
| H-80 | 60.1 | 65.2 | **70.9** | 68.5 | 68.1 | 68.6 | 69.0 |
| H-40 | 57.2 | 43.3 | **64.9** | 59.6 | 57.3 | 58.3 | 56.7 |
| **GSM8K - SCA ($\alpha = 10$)** | | | | | | | |
| SCA-100 | 23.7 | **30.1** | 26.8 | 27.4 | 27.5 | 27.0 | 27.5 |
| SCA-80 | 23.3 | **28.1** | 26.2 | 26.7 | 26.6 | 26.0 | 26.6 |
| SCA-40 | 16.6 | 15.3 | **18.0** | 17.9 | 16.8 | 16.5 | 16.7 |
| **SVAMP - SCA ($\alpha = 10$)** | | | | | | | |
| SCA-100 | 46.7 | 43.3 | **51.3** | 49.4 | 43.1 | 48.9 | 45.4 |
| SCA-80 | 46.6 | 41.2 | **50.7** | 48.1 | 41.6 | 47.7 | 44.9 |
| SCA-40 | 40.6 | 27.8 | **42.7** | 40.8 | 34.8 | 38.0 | 35.8 |
| **ASDIV - SCA ($\alpha = 10$)** | | | | | | | |
| SCA-100 | **84.4** | 67.3 | 71.2 | 69.3 | 68.9 | 69.9 | 70.4 |
| SCA-80) | **81.1** | 66.1 | 71.0 | 68.8 | 68.5 | 69.2 | 69.5 |
| SCA-40 | 66.0 | 50.4 | **67.3** | 63.3 | 61.5 | 61.9 | 61.2 |
| **GSM8K - CCA ($\alpha = 10, \beta = 40$)** | | | | | | | |
| CCA-100 | 23.7 | **30.1** | 26.8 | 27.4 | 27.5 | 27.0 | 27.5 |
| CCA-80 | 23.3 | **28.1** | 26.2 | 26.7 | 26.6 | 26.0 | 26.6 |
| CCA-40 | 16.6 | 15.3 | **18.0** | 17.9 | 16.8 | 16.5 | 16.7 |
| **SVAMP - CCA ($\alpha = 10, \beta = 40$)** | | | | | | | |
| CCA-100 | 46.7 | 43.3 | **51.3** | 49.4 | 43.1 | 48.9 | 45.4 |
| CCA-80 | 46.6 | 41.2 | **50.7** | 48.1 | 41.6 | 47.7 | 44.9 |
| CCA-40 | 40.6 | 27.8 | **42.7** | 40.8 | 34.8 | 38.0 | 35.8 |
| **ASDIV - CCA ($\alpha = 10, \beta = 40$)** | | | | | | | |
| CCA-100 | **84.4** | 67.3 | 71.2 | 69.3 | 68.9 | 69.9 | 70.4 |
| CCA-80 | **81.1** | 66.1 | 71.0 | 68.8 | 68.5 | 69.2 | 69.5 |
| CCA-40 | **66.0** | 50.4 | 67.3 | 63.3 | 61.5 | 61.9 | 61.2 |

Table 2: Falcon-40b

### 7.4 Understanding the effect of CCoT

While the previous experiments demonstrate that CCoT achieves a clear reduction in the number of words produced, thus improving the trade-off between accuracy and generation time. The scores introduced in Section 6 are used in the following analysis to quantify the benefits in terms of conciseness (RQ3). As detailed in Section 6, the conciseness of an answer is assumed to depend on two key properties: *(i)* the number of steps required to produce a response, and *(ii)* assuming a similar number of steps, the amount of information repeated across those steps.

**Number of Steps.** To evaluate the first property, we plotted in Figure 4 the distribution of the percentage of generated answers based on the *number of reasoning steps* (Fu et al., 2022) for Llama2-70b (top row) and Falcon-40b (bottom row) across the three datasets. The results reveal a significantly different distribution in the number of steps across all datasets with Falcon-40b and for the ASDIV dataset with Llama2-70b. In fact, in these cases, the CCoT curves are clearly concentrated toward a reduced number of steps compared to CoT, indicating that the CCoT prompt consistently generates responses with fewer steps.

While the analysis of the number of steps explains the improved conciseness achieved with CCoT, more detailed analysis of the redudancy and information flow are required to understand the enhanced conciseness observed with Llama2-70b on GSM8K and SVAMP, where the number of steps follows a trend similar to that of CoT.

**Redundancy Evaluation.** As discussed in the previous paragraph, the distribution of reasoning steps for Llama2-70b on the GSM8K and SVAMP datasets (Figures 4a and 4b) reveals that the quartile ranges of the answer distributions for CCoT and CoT answers are very similar. Specifically, in both datasets, the interquartile range spans approximately 4 to 12 steps (Q1 to Q3) for SVAMP and around 5 to 14 steps for GSM8K. This suggests comparable step counts across CCoT and CoT answers, within such ranges of reasoning steps.

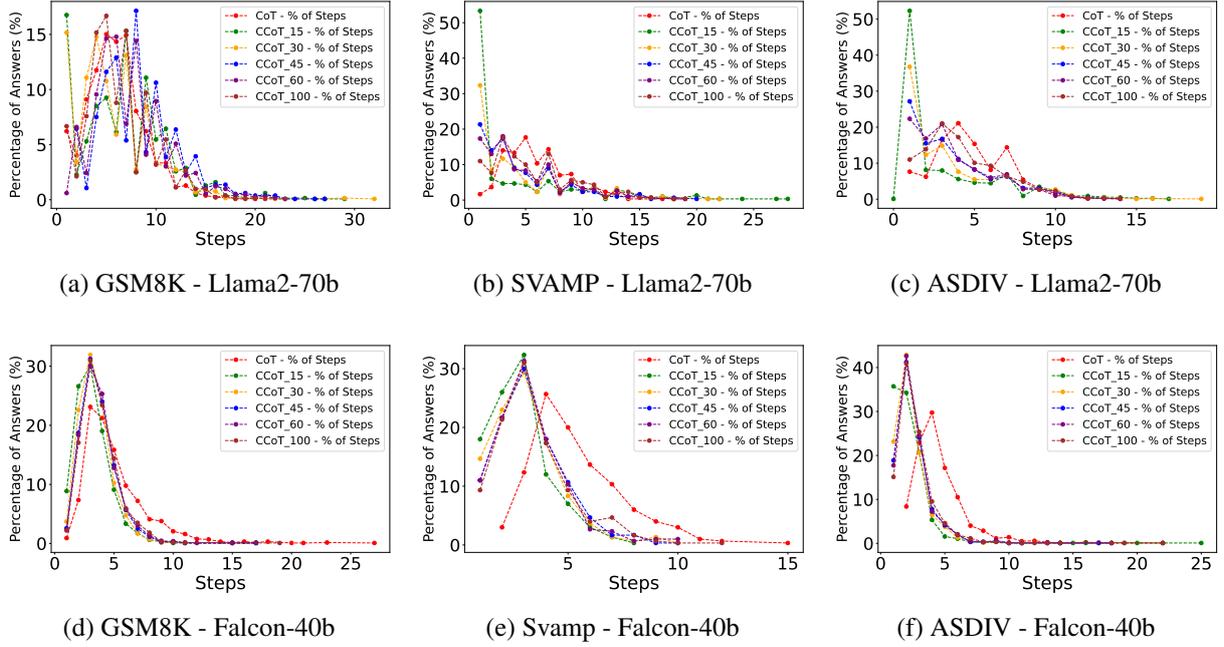

Figure 4: Percentage of number of generated answers as per 'no. of reasoning steps' of Llama2-70b (top row) and with Falcon-40b (bottom row) on the GSM8K (a), SVAMP (b), and ASDIV (c) test datasets.

To evaluate the conciseness of these reasoning steps more effectively, we present redundancy scores in Figure 5. This figure highlights the differences in redundancy between CCoT and CoT answers, categorized by their number of steps. Notably, when focusing on step intervals within the interquartile range (Q1 to Q3, where most answers fall), which are highlighted in grey in the plots, redundancy scores are consistently higher for CoT than for CCoT. This demonstrates that CCoT achieves improved syntactic conciseness compared to CoT, even when the number of steps is similar.

To better quantify the reduction in redundancy achieved by CCoT, we define the *Mean Redundancy Reduction (MRR)* across single-step redundancies $RMS_i$ for all reasoning steps $i$, and the *Overall Redundancy Reduction (ORR)* as:

$$\text{MRR} = \frac{1}{n} \sum_{i=1}^{n} \left( \frac{\text{CoT RMS}_i - \text{CCoT RMS}_i}{\text{CoT RMS}_i} \times 100 \right)$$

$$\text{ORR} = \frac{\text{CoT RMS} - \text{CCoT RMS}}{\text{CoT RMS}} \times 100$$

Table 3 presents the average ORR and MRR calculated for the SVAMP and GSM8K datasets, focusing on steps within the interquartile range (Q1 to Q3). The results indicate that CCoT consistently reduces redundancy. For SVAMP, the redundancy reduction ranges from 19.77% to 22.72% (ORR), while for GSM8K, it ranges from 12.64% to 24.74% (ORR).

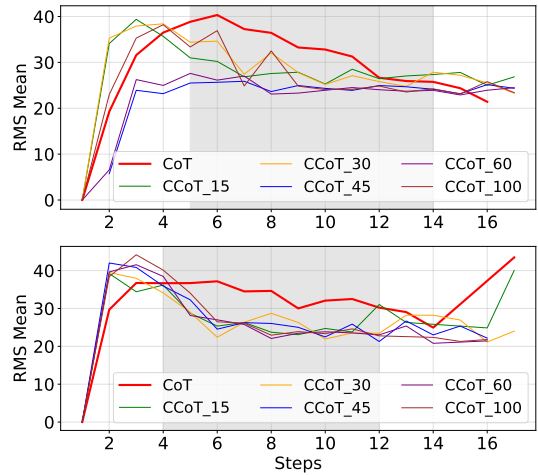

Figure 5: RMS score (lower values indicate better conciseness) of Llama2-70b on the GSM8K (top) and SVAMP (bottom) datasets. The grey area highlights the portion of the answer distribution between the Q1 and Q3 quartiles.

| Method | Dataset | CCoT15 | CCoT30 | CCoT45 | CCoT60 | CCoT100 |
|---|---|---|---|---|---|---|
| MRR | SVAMP | 20.02 | 22.50 | 21.48 | 22.92 | 20.14 |
|  | GSM8K | 13.58 | 11.65 | 22.81 | 23.23 | 16.33 |
| ORR | SVAMP | 20.22 | 22.57 | 21.29 | 22.72 | 19.77 |
|  | GSM8K | 15.34 | 12.64 | 24.62 | 24.74 | 16.81 |

Table 3: Mean and Overall Redundancy Reduction for SVAMP and GSM8K Datasets

**Information Flow Evaluation.** To demonstrate an improved level of conciseness also from a semantic perspective, we analyze the information flow. Specifically, the median number of steps in the answer distributions for CoT and CCoT (Figures 4a and 4b) is approximately 8 for both the SVAMP and GSM8K datasets. Thus, focusing on answers with a total of 8 steps, we present in Tables 4 the *Information Flow* between consecutive steps $i \to j$ for Llama2-70b on the GSM8K (top) and SVAMP (bottom) datasets.

In Table 4, CCoT-15 exhibits the largest reductions across all steps, for instance, ranging from 26% to 41% compared to CoT scores for GSM8K, while CCoT-100 shows the smallest reductions (4% to 20%), preserving more redundant semantic information. Generally, the middle steps show the largest differences, especially for aggressive variations like CCoT-15 and CCoT-45, whereas early and late steps retain more information flow. In summary, a lower information flow indicates that the model effectively retains, step by step, only the logically necessary information required to arrive at a correct answer.

| GSM8K - Llama2-70b | | | | | | |
|---|---|---|---|---|---|---|
| Steps | CoT | CCoT-15 | CCoT-30 | CCoT-45 | CCoT-60 | CCoT-100 |
| 1 ⇒ 2 | 0.5287 | 0.3417 | 0.39 | 0.49 | 0.49 | 0.43 |
| 2 ⇒ 3 | 0.59 | 0.39 | 0.45 | 0.31 | 0.32 | 0.53 |
| 3 ⇒ 4 | 0.56 | 0.37 | 0.43 | 0.36 | 0.37 | 0.47 |
| 4 ⇒ 5 | 0.55 | 0.38 | 0.41 | 0.36 | 0.36 | 0.46 |
| 5 ⇒ 6 | 0.56 | 0.39 | 0.42 | 0.36 | 0.37 | 0.51 |
| 6 ⇒ 7 | 0.56 | 0.42 | 0.45 | 0.37 | 0.37 | 0.54 |
| 7 ⇒ 8 | 0.52 | 0.42 | 0.47 | 0.57 | 0.56 | 0.50 |
| SVAMP - Llama2-70b | | | | | | |
| Steps | CoT | CCoT-15 | CCoT-30 | CCoT-45 | CCoT-60 | CCoT-100 |
| 1 ⇒ 2 | 0.54 | 0.32 | 0.41 | 0.41 | 0.30 | 0.32 |
| 2 ⇒ 3 | 0.52 | 0.35 | 0.47 | 0.43 | 0.35 | 0.35 |
| 3 ⇒ 4 | 0.51 | 0.34 | 0.46 | 0.40 | 0.30 | 0.37 |
| 4 ⇒ 5 | 0.51 | 0.37 | 0.43 | 0.39 | 0.30 | 0.30 |
| 5 ⇒ 6 | 0.51 | 0.38 | 0.41 | 0.44 | 0.36 | 0.37 |
| 6 ⇒ 7 | 0.50 | 0.41 | 0.48 | 0.40 | 0.36 | 0.40 |
| 7 ⇒ 8 | 0.47 | 0.35 | 0.43 | 0.37 | 0.34 | 0.50 |

Table 4: Information Flow Mean Values comparison for GSM8K (top) and SVAMP (bottom) across answers with 8 steps. Better information flow indicates lower semantic conciseness between steps (highlighted with blue-like colors in the tables).

### 7.5 Ability to control the output length

The previous experiments looked at how CCoT strategies can affect the accuracy and generation time in the average. However, despite the discussed benefits, it is also crucial to understand how CCoT prompting can effectively limit the output length for each addressed sample (RQ3). This can be useful for better tuning the length parameter in the CCoT prompt or identifying the conditions in which the proposed prompting strategy fails to compress the output. To evaluate the ability of an LLM to produce concise answers in response to a given prompting, we analyzed in Figure 6 the output length under different CCoT length constraints.

Figure 6 shows the statistics on the length of the answers provided by addressed models with the GSM8K test set. Each box plot represents the output lengths between the 5th and the 95th percentiles of all tested samples, the blue line represents the provided CCoT length constraint, the red line denotes the median, while the greed dot the mean. Ideally, a model respecting the given length constraint for each tested sample should have the entire distribution below the blue line.

As depicted in Figure 6, CoT based LLMs tend to produce long answers if not explicitly constrained, significantly impacting the generation time. The imposed length constraint in the CCoT prompt significantly affects the output length, although in practice LLMs are not always able to respect the given limit, especially for smaller values, such as 15, 30, or 40, which are more challenging.

To summarize, given the nature of the CCoT prompting, it is reasonable to consider a tolerance margin in respecting the requested length. To this end, in the following paragraphs we evaluate the considered models by the metrics proposed in Section 4, which extend the accuracy by also accounting for conciseness.

## 8 Final Remark and Conclusion

**Limitations and Future Directions.** From the findings revealed by the conducted experiments, a key insight is that for sufficient-scale models, such as Falcon-40b and Llama2-70b, CCoT effectively achieves a better trade-off between accuracy and efficiency. However, we acknowledge that smaller models may struggle to improve this trade-off, often producing incorrect answers when attempting to limit reasoning length. A detailed analysis of these findings is provided in the appendix B. We believe this aligns with the inherent capability of LLMs to exhibit a sense of understanding of output length in their generated responses(Bhargava et al., 2023). Future investigations should explore addressing this constraint and proposed metrics not only through inference prompts but also as part of a fine-tuning strategy, which could prove beneficial even for smaller-sized models.

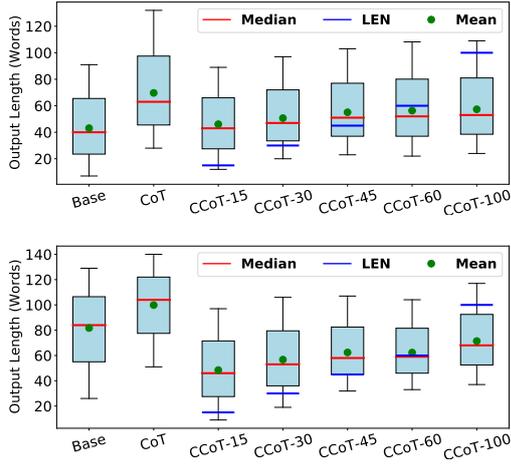

Figure 6: Distribution of output lengths (y-axis) between the 5th and 95th percentiles for different models and prompting strategies using the GSM8K test set. The top plot shows the results for Falcon-40b, while the bottom plot presents those for Llama2-70b.

Another point involves a deeper analysis of other potential benefits of conciseness in LLMs, beyond a study of the efficiency, which is the main scope of this work. For instance, while most analyses justify conciseness through a reduction in the number of steps (see Section 7.4), a detailed examination of redundancy and information reveals that this behavior also emerges in scenarios where CCoT and CoT have a similar number of steps. We believe that these metrics can be further leveraged to explore additional benefits of concise generation, such as mitigating hallucinations or reducing error propagation. For instance, a lower Information Flow score suggests that the model retains only the logically necessary information required to reach a correct answer, step by step, while excluding superfluous details. This approach could help reduce the risk of error propagation by filtering out unnecessary or irrelevant information at each step (Li et al., 2024).

**Conclusion.** This work explored the importance of conciseness in answers generated by LLMs for text-to-text tasks, introducing three new metrics to evaluate both conciseness and correctness based on user-defined parameters. Additionally, it proposed a prompt engineering approach, Constrained Chain-of-Thought, to get a better control of output length of generated answer ,thus impacting the inference time. Furthermore, it deepened the analysis of conciseness by evaluating its significance within the generated reasoning steps, particularly in terms of redundancy and information flow.

Experimental results demonstrated how the proposed Constrained CoT (CCoT) effectively highlights the trade-off (RQ1) and presents a unified perspective through new metrics (RQ2), particularly when applied to large LLMs such as Falcon-40b and Llama2-70b. Furthermore, we provide an in-depth analysis of the concept of conciseness, the capability to control output length, and the importance of information within the generated reasoning steps (RQ3).

In conclusion, this work emphasizes the need to focus on the conciseness of LLMs by proposing novel performance metrics that evaluate both the correctness of the output and its length. Additionally, the proposed CCoT prompting offers a simple yet effective strategy to enhance conciseness, potentially paving the way for new research directions to make LLMs more predictable and efficient.

# Appendix of the paper "Concise Thoughts: Impact of Output Length on LLM Reasoning and Cost"

# Appendices

## A  Additional analysis of the conciseness

### A.1  Analysis of the RMS

Figures 7, 8, and 9 present the RMS values for Llama2-70b and Falcon-40b across the three datasets analyzed in the paper. These results specifically highlight scenarios where the model successfully answers questions with the CCoT prompt but not with the CoT prompt. The purpose of these plots is to check whether the behavior of CCoT in terms of redundancy is coherent with that observed in Section 7.4. In addition, these plots help to investigate the potential correlation between the accuracy of the model's responses and the reduction in redundancy. The plots in the left column correspond to Llama2-70b, while those in the right column correspond to Falcon-40b. In each plot, we also highlight the interval between Q1 and Q3 for CoT and CCoT answers, along with the overlap range of these intervals.

As we can see from these plots, the redundancy behavior for CCoT and CoT closely resembles the trend observed in Section 7.4 calculated across all questions. This suggests that the reduction in redundancy (for Llama2-70b) and number of steps (for Falcon-40b) achieved by CCoT is consistent and provides more accurate and concise answers compared to CoT. In fact, the reduction in terms of redundancy or number of steps in the cases where CoT fails indicates that the CCoT prompt provides better guidance to the model, even in complex or ambiguous scenarios where CoT tends to struggle.

### A.2  Information Flow

Since the main paper examined the information flow in answers with 8 steps based on the CCoT answer distribution, we further demonstrate the effectiveness of CCoT by analyzing step-by-step information flow for another median step count, specifically 9 steps. Tables 5 and 6 present the scores for Llama2-70b on GSM8K and SVAMP, respectively.

In Table 5, for GSM8K, all steps show a reduced repetition of semantic information for CCoT, ex-cept for CCoT-45 at the initial step and CCoT-100 at the final step. A different behavior is observed at the final step for all CCoTs in the other dataset (SVAMP), as shown in Table 6, although semantic information is retained throughout the steps.

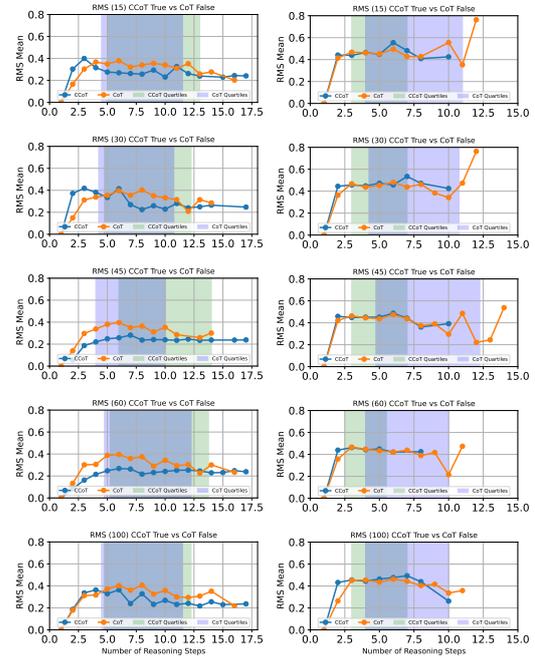

Figure 7: RMS mean score comparison between CCoT-true and CoT-false on the GSM8K dataset, with Llama2-70b (left side) and Falcon-40b (right side). The overlapped highlighted area illustrates the answer distribution between Q1 and Q3.

We also calculated the semantic information flow for answers with 8 and 9 steps, based on the median step distribution of CCoT answers generated by Falcon-40b. The results are presented in Tables 7 and 8 for the GSM8K and SVAMP datasets, respectively. Interestingly, Falcon-40b exhibits contrasting behavior in terms of information scores, often displaying higher values, which suggest greater repetition of semantic information across steps. This behavior is likely influenced by its medium-scale architecture and the nature of its training dataset, which may not be well-suited for handling constrained reasoning tasks.

However, as also highlighted in the main text of the paper, it is important to note that Falcon-40b achieves improved conciseness due to the significantly lower average number of steps it produces. This allows for quicker reasoning decisions in many samples. This observation clarifies that the analysis reported in Tables 7 and 8 is inherently unbalanced, as the number of Falcon-40b answers with 8 and 9 steps under CCoT is smaller than those under CoT. Please note that in Table 8, there are no information scores for CCoT-15 because Falcon-40b does not generate answers with 8 or 9 steps when constrained to a reasoning length of up to 15 tokens.

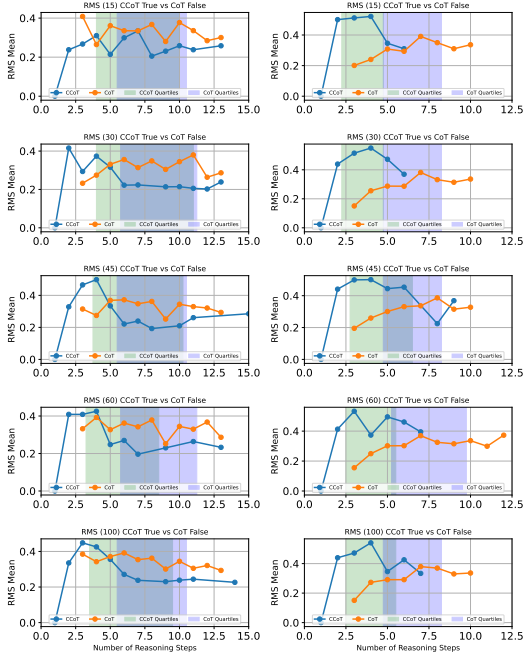

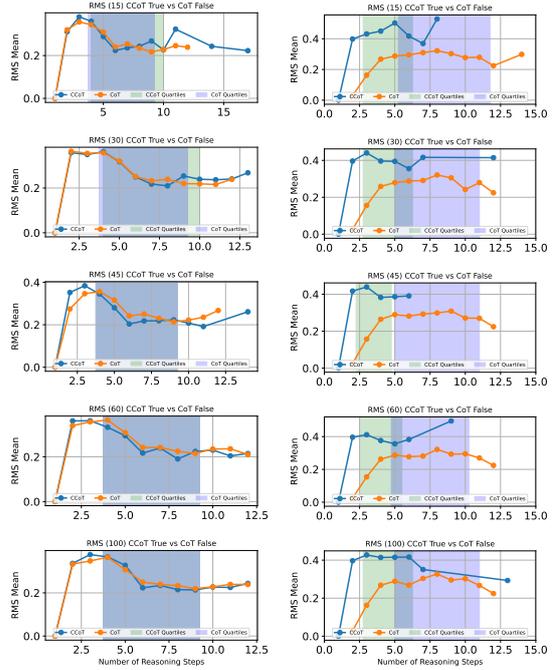

Figure 8: RMS mean score comparison between CCoT-true and CoT-false on the SVAMP dataset, with Llama2-70b (left side) and Falcon-40b (right side). The overlapped highlighted area illustrates the answer distribution between Q1 and Q3.

Figure 9: RMS mean score comparison between CCoT-true and CoT-false on the ASDIV dataset, with Llama2-70b shown (left side) and Falcon-40b (right side). The overlapped highlighted area illustrates the answer distribution between Q1 and Q3.

| Steps | CoT | CCoT-15 | CCoT-30 | CCoT-45 | CCoT-60 | CCoT-100 |
|---|---|---|---|---|---|---|
| 1 ⇒ 2 | 0.48 | 0.29 | 0.32 | 0.48 | 0.45 | 0.33 |
| 2 ⇒ 3 | 0.54 | 0.38 | 0.40 | 0.35 | 0.33 | 0.36 |
| 3 ⇒ 4 | 0.49 | 0.36 | 0.37 | 0.40 | 0.38 | 0.35 |
| 4 ⇒ 5 | 0.49 | 0.35 | 0.36 | 0.38 | 0.37 | 0.34 |
| 5 ⇒ 6 | 0.51 | 0.38 | 0.40 | 0.39 | 0.38 | 0.36 |
| 6 ⇒ 7 | 0.49 | 0.37 | 0.37 | 0.39 | 0.36 | 0.36 |
| 7 ⇒ 8 | 0.50 | 0.36 | 0.37 | 0.43 | 0.40 | 0.35 |
| 8 ⇒ 9 | 0.55 | 0.50 | 0.54 | 0.45 | 0.51 | 0.61 |

Table 5: GSM8K Llama2-70b Information Flow Mean Values comparison across answers with 9 setps

| Steps | CoT | CCoT-15 | CCoT-30 | CCoT-45 | CCoT-60 | CCoT-100 |
|---|---|---|---|---|---|---|
| 1 ⇒ 2 | 0.46 | 0.29 | 0.32 | 0.30 | 0.29 | 0.32 |
| 2 ⇒ 3 | 0.49 | 0.31 | 0.36 | 0.33 | 0.34 | 0.34 |
| 3 ⇒ 4 | 0.46 | 0.33 | 0.34 | 0.31 | 0.30 | 0.32 |
| 4 ⇒ 5 | 0.45 | 0.32 | 0.34 | 0.32 | 0.29 | 0.31 |
| 5 ⇒ 6 | 0.49 | 0.32 | 0.39 | 0.37 | 0.36 | 0.34 |
| 6 ⇒ 7 | 0.51 | 0.31 | 0.37 | 0.41 | 0.34 | 0.33 |
| 7 ⇒ 8 | 0.47 | 0.32 | 0.38 | 0.38 | 0.37 | 0.39 |
| 8 ⇒ 9 | 0.49 | 0.54 | 0.63 | 0.68 | 0.62 | 0.58 |

Table 6: SVAMP Llama70b Information Flow Mean Values comparison across answers with 9 setps

## B  Testing CCoT with smaller LLMs.

In this experiments, we investigate the capability of larger set of LLMs to handle the CCoT prompting. Specifically, in Figure 10, we present an evaluation conducted on five LLMs, including small and medium-sized models such as Falcon-7b (Almazrouei et al., 2023) and LLama2-7b (Touvron et al., 2023). The results acknowledge some difficulties in addressing the CCoT prompt when considering smaller models.

We believe that such different outcomes of CCoT prompting can be attributed to various factors, such as the training data, the approach used to train the model, the model size, and the technique adopted during training. For instance, Llama2-70b is an autoregressive large-scale language model fine-tuned with human feedback, trained on a diverse combination of generic and open-source datasets. Such technical measures contribute to making CCoT effective in controlling the output length while improving the model accuracy. The Falcon-40b model, in contrast, is smaller than Llama2-70b and trained

Table 7: GSM8K Falcon-40b Information Flow Mean Values Comparison

| Steps | CoT | CCoT-15 | CCoT-30 | CCoT-45 | CCoT-60 | CCoT-100 |
|---|---|---|---|---|---|---|
| $1 \Rightarrow 2$ | 0.58 | 0.65 | 0.64 | 0.65 | 0.61 | 0.59 |
| $2 \Rightarrow 3$ | 0.62 | 0.69 | 0.68 | 0.70 | 0.68 | 0.64 |
| $3 \Rightarrow 4$ | 0.60 | 0.74 | 0.76 | 0.74 | 0.68 | 0.63 |
| $4 \Rightarrow 5$ | 0.59 | 0.66 | 0.63 | 0.64 | 0.64 | 0.66 |
| $5 \Rightarrow 6$ | 0.61 | 0.62 | 0.72 | 0.67 | 0.66 | 0.64 |
| $6 \Rightarrow 7$ | 0.60 | 0.61 | 0.66 | 0.68 | 0.68 | 0.62 |
| $7 \Rightarrow 8$ | 0.59 | 0.50 | 0.55 | 0.54 | 0.56 | 0.59 |

(a) Answers with 8 steps

| Steps | CoT | CCoT-15 | CCoT-30 | CCoT-45 | CCoT-60 | CCoT-100 |
|---|---|---|---|---|---|---|
| $1 \Rightarrow 2$ | 0.55 | 0.62 | 0.69 | 0.61 | 0.58 | 0.67 |
| $2 \Rightarrow 3$ | 0.58 | 0.62 | 0.65 | 0.65 | 0.57 | 0.73 |
| $3 \Rightarrow 4$ | 0.54 | 0.62 | 0.67 | 0.63 | 0.61 | 0.73 |
| $4 \Rightarrow 5$ | 0.54 | 0.62 | 0.64 | 0.46 | 0.55 | 0.63 |
| $5 \Rightarrow 6$ | 0.54 | 0.69 | 0.74 | 0.67 | 0.64 | 0.76 |
| $6 \Rightarrow 7$ | 0.55 | 0.70 | 0.59 | 0.48 | 0.56 | 0.66 |
| $7 \Rightarrow 8$ | 0.55 | 0.65 | 0.77 | 0.68 | 0.65 | 0.75 |
| $8 \Rightarrow 9$ | 0.59 | 0.52 | 0.60 | 0.46 | 0.48 | 0.68 |

(b) Answers with 9 steps

Table 8: SVAMP Falcon-40b Information Flow Mean Values Comparison

| Steps | CoT | CCoT-15 | CCoT-30 | CCoT-45 | CCoT-60 | CCoT-100 |
|---|---|---|---|---|---|---|
| $1 \Rightarrow 2$ | 0.26 | 0.60 | 0.58 | 0.55 | 0.58 | 0.61 |
| $2 \Rightarrow 3$ | 0.68 | 0.58 | 0.53 | 0.53 | 0.60 | 0.52 |
| $3 \Rightarrow 4$ | 0.60 | 0.68 | 0.52 | 0.60 | 0.49 | 0.54 |
| $4 \Rightarrow 5$ | 0.61 | 0.56 | 0.51 | 0.56 | 0.63 | 0.60 |
| $5 \Rightarrow 6$ | 0.63 | 0.51 | 0.65 | 0.55 | 0.68 | 0.64 |
| $6 \Rightarrow 7$ | 0.63 | 0.95 | 0.76 | 0.52 | 0.64 | 0.62 |
| $7 \Rightarrow 8$ | 0.55 | 0.73 | 0.57 | 0.60 | 0.56 | 0.65 |

(a) Answers with 8 Steps

| Steps | CoT | CCoT-15 | CCoT-30 | CCoT-45 | CCoT-60 | CCoT-100 |
|---|---|---|---|---|---|---|
| $1 \Rightarrow 2$ | 0.26 | - | 0.64 | 0.62 | 0.56 | 0.61 |
| $2 \Rightarrow 3$ | 0.65 | - | 0.59 | 0.55 | 0.53 | 0.66 |
| $3 \Rightarrow 4$ | 0.57 | - | 0.66 | 0.52 | 0.62 | 0.63 |
| $4 \Rightarrow 5$ | 0.58 | - | 0.74 | 0.67 | 0.55 | 0.54 |
| $5 \Rightarrow 6$ | 0.60 | - | 0.67 | 0.47 | 0.67 | 0.61 |
| $6 \Rightarrow 7$ | 0.57 | - | 0.86 | 0.47 | 0.75 | 0.64 |
| $7 \Rightarrow 8$ | 0.63 | - | 0.77 | 0.69 | 0.63 | 0.62 |
| $8 \Rightarrow 9$ | 0.54 | - | 0.61 | 0.56 | 0.57 | 0.43 |

(b) Answers with 9 Steps

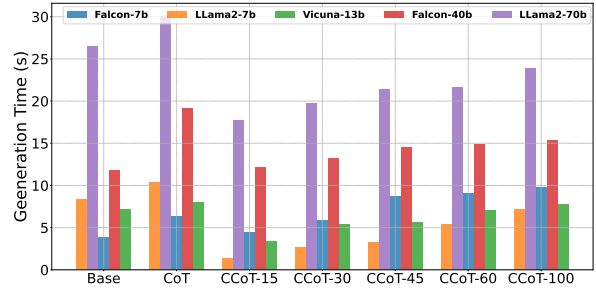

(a) Generation time

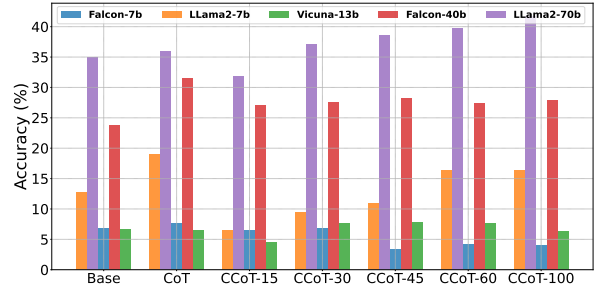

(b) Accuracy

Figure 10: Generation time (a) and accuracy (b) of five LLMs (Llama2-7b, Llama2-70b, Falcon-7b, Falcon-40b, and Vicuna-13b) on the GSM8K test dataset. Each model is evaluated using plain promt (base), CoT, and CCoT with different length constraints.

also for smaller-sized models.

on a different dataset (the dedicated RefinedWeb data (Penedo et al., 2023)). While CCoT does not improve the accuracy of the model with respect to CoT, it still performs better than the base plain prompting, offering a trade-off by reducing generation times compared to CoT. Vicuna-13b also provides competitive results across different prompts, as it is a fine-tuned version of Llama2 and smaller than the previous Llama2-70b.

Conversely, small-scale LLMs, such as Falcon-7b and Llama2-7b, are not capable of properly handling the constrained prompting conditions in CCoT, resulting in higher generation times (as shown for Falcon-7b with large length values in CCoT) or incorrect answers with short CCoT values in Llama2-7b. This suggests that model size and training strategies severely impact the effectiveness of CCoT.

Considering the observations presented here, we believe that future directions could face potential training and fine-tuning strategy to integrate a better awareness and capability fo thandlign lengths